\documentclass[11pt]{article}

\usepackage[preprint]{acl}

\usepackage{times}
\usepackage{latexsym}
\usepackage[T1]{fontenc}
\usepackage[utf8]{inputenc}
\usepackage{microtype}
\usepackage{inconsolata}
\usepackage{graphicx}
\usepackage{booktabs}
\usepackage{multirow}
\usepackage{amsmath}
\usepackage{xcolor}
\usepackage{enumitem}
\usepackage{float}
\setlist{nosep,leftmargin=*}

\newcommand{\vrand}{\textsc{random}}
\newcommand{\vrel}{\textsc{rel-uncited}}
\newcommand{\vcite}{\textsc{cited-other}}
\newcommand{\planted}[1]{\colorbox{gray!22}{\textbf{#1}}}

\title{Attributable Post-Rationalization in RAG Citations:\\A Controlled Reproduction and an RLVR Comparison}

\author{Mehedi Khan \and Md. Shariful Islam Bhuyan \\
  Department of Computer Science and Engineering \\
  Bangladesh University of Engineering and Technology \\
  \texttt{mehedi.72.khan@gmail.com}, \texttt{sharifulislam@cse.buet.ac.bd}}

\begin{document}
\maketitle

\begin{abstract}
A RAG system can hand you the right answer and cite a source it did not actually use. Models output these unfaithful citations via \emph{post-rationalization}: they write the answer first and then attach a citation to whatever passage looks close enough.
Search agents are now trained with reinforcement learning from verifiable rewards (RLVR), which pays them for getting the answer right. We asked whether that training also teaches them to cite honestly. 

Improving an existing methodology with a required control, we compared an instruction-tuned model against three RLVR agents trained from it, on four question-answering datasets, using only free-tier Kaggle GPUs. Post-rationalization is everywhere: on Wikipedia-based questions roughly one citation in seven is unfaithful. RLVR does not fix it. The agents post-rationalize at their base model's rate, and one lands slightly worse. Rewarding correct answers buys nothing in citation faithfulness, so faithfulness has to be trained and measured on its own terms.
\end{abstract}

\section{Introduction}
\label{sec:intro}

\begin{figure}[t]
  \centering
  \small
  \fbox{%
  \begin{minipage}{0.94\columnwidth}
    \textbf{Question:} Who plays Joker in Batman:  The Dark Knight?

    \vspace{4pt}
    \textbf{Answer (clean context):}\\
    Heath Ledger plays the Joker in Batman: The Dark Knight [1].

    \vspace{4pt}
    \hrule
    \vspace{4pt}
    \textbf{Passage [1]} \emph{(Heath Ledger)}: \ldots\ In his penultimate film performance, Ledger played the Joker in Christopher Nolan's 2008 film ``The Dark Knight'' \ldots

    \vspace{4pt}
    \textbf{Passage [4]} \emph{(Christmas with the Joker)}: \ldots\ Batman leaps back down to the ground, telling Robin to keep an eye out for the Joker. \ldots\ \planted{Heath Ledger plays the}

    \vspace{4pt}
    \hrule
    \vspace{4pt}
    \textbf{Answer (after planting):}\\
    Heath Ledger plays the Joker in Batman The Dark Knight [4].
  \end{minipage}}
  \caption{A real case from our baseline model on NQ. Passage [4] is a plot summary of a Batman TV episode; it never mentions Heath Ledger or the film. We appended four words from the model's own answer to it (shaded). The model then cites [4] instead of [1], although [1] is still in the context and supports the claim. Passages are abridged, and the other three retrieved passages are not shown.}
  \label{fig:example}
\end{figure}

Citations are the part of a RAG answer that a careful reader actually checks. The system retrieves passages, writes an answer, and marks each sentence with the passage it came from \citep{lewis2021rag,gao-etal-2023-enabling}; the marker is a promise about provenance. Follow [1], the promise says, and you will find where this sentence came from.

Models break that promise more often than they look like they do. \citet{wallat-etal-2025-correctness} showed that a model will cite a passage it never used, a behavior they call \emph{post-rationalization}: the answer comes first, and the citation gets attached afterward to whatever passage looks like a plausible source. What makes this failure slippery is that nothing on the surface goes wrong. The answer is right. The citation is present, well formed, and points at a real passage. Only someone who reads that passage closely notices that it never supported the claim.

Figure~\ref{fig:example} shows how thin the connection can get. Our baseline model correctly names Heath Ledger as the Joker and cites the Heath Ledger article. We then copied four words of the model's own sentence into a plot summary of a Batman cartoon and asked again. The answer did not change. The citation moved to the cartoon passage, which never answers the original question, while the Ledger article stayed right there in the context. Four words were enough to make the model cite a source it did not use during inference.

This gets more pressing as RAG systems move from prompting to training. Search agents such as Search-R1 \citep{jin2025searchr1}, ReSearch \citep{chen2025research}, and R-Search \citep{zhao2025rsearch} learn to search, read, and answer through reinforcement learning from verifiable rewards (RLVR), where the reward comes from checking the final answer. These agents get better at finding answers. Whether they also get better at pointing to the passage they used is a separate question, and their reward never asks it. That is the question we set out to answer.


With a measurement we trust, we compared four models that share one base, Qwen2.5-7B-Instruct \citep{qwen2025technical}: the instruction-tuned model itself and three open-weight RLVR agents trained from it. One shared base is what makes the comparison clean, since any gap between an agent and the baseline comes from training rather than from a different model family. We ran all four across four question-answering datasets on Kaggle's T4 GPUs.

Two findings stand out. Post-rationalization is common and it is not an artifact of RL: our untrained baseline cites unfaithfully for roughly one out of seven questions on the Wikipedia-based datasets. Under fixed-context inference, RLVR leaves it untouched. Search-R1 and ReSearch track their base model within a couple of points in either direction, and R-Search lands slightly worse rather than better. Our sample sizes would have caught any shift of about seven percentage points, so a large improvement is ruled out. Training a model to answer correctly, it turns out, teaches it nothing about citing honestly, which means citation faithfulness has to be rewarded and measured directly.

We contribute:
\begin{itemize}
  \item \textbf{A no-planting control as an improvement to the existing methodology}, to get a more accurate idea about post-rationalization.
  \item \textbf{The first controlled measurement of post-rationalization in RLVR search agents}, holding the base model, retrieval corpus, prompt, and scorer fixed across four datasets.
  \item \textbf{Evidence that answer-only rewards leave citation faithfulness exactly where they found it}, with a stated detection bound, plus a concrete proposal for folding attribution into the reward.
  \item \textbf{A full reproduction on free-tier hardware}, replacing the 104B Command R+ model of the original study with an open 7B model on two T4s.
\end{itemize}

\section{Related Work}
\label{sec:related}

\paragraph{Citations in RAG.} RAG conditions generation on retrieved passages \citep{lewis2021rag}, and benchmarks such as ALCE ask models to cite those passages sentence by sentence \citep{gao-etal-2023-enabling}. Most evaluations score whether a citation is \emph{correct}: does the cited passage support the sentence? \citet{wallat-etal-2025-correctness} pushed on a harder question: did the model actually \emph{use} it? Those two can come apart. A model that already knows the answer can write it down, scan the passages, and attach the one that fits, producing a citation that is correct and unfaithful at the same time. Their planting probe exposes the gap, and we reproduce it, then add the control it was missing (Section~\ref{sec:method}).

\paragraph{RLVR search agents.} Search-R1 \citep{jin2025searchr1}, ReSearch \citep{chen2025research}, and R-Search \citep{zhao2025rsearch} teach a language model to interleave search calls with reasoning, rewarding it when the final answer checks out against a reference. All three release open weights trained from Qwen2.5-7B-Instruct, which lets us line each agent up against the exact model it grew from. Their papers report answer quality. None of them reports how many citations are faithful, which is the gap we fill.

\paragraph{Faithful reasoning is a different question.} VERITAS \citep{xu2026veritas} showed that the reasoning traces of Search-R1 and ReSearch do not always line up with the searches they issue, the information they retrieve, or the answers they produce. That is faithfulness \emph{within} the reasoning chain. We ask about faithfulness \emph{to the evidence}: is the cited passage the one the answer leaned on? An agent can narrate its search honestly and still misattribute the result, or reason sloppily and cite correctly. The two findings sit side by side rather than in conflict.

\section{Measuring Post-Rationalization}
\label{sec:method}

\subsection{The planting probe}
\label{sec:method-probe}

We follow the probe of \citet{wallat-etal-2025-correctness}, which runs in three steps.

\begin{enumerate}
  \item \textbf{Answer.} The model answers a question from five retrieved passages. Every sentence ends with exactly one citation, so each sentence and its marker form one \emph{claim}.
  \item \textbf{Plant.} We lift a short span out of the claim and append it to a passage the model did \emph{not} cite for that claim, then re-run the same model on the doctored context with the same prompt.
  \item \textbf{Check.} We ask whether the model now cites the doctored passage for that claim.
\end{enumerate}

The logic is a swap test. The doctored passage shares nothing with the claim except the words we pasted into it, and the passage the model originally cited never leaves the context. A model that reads for meaning has no reason to move. A model that matches surface strings has every reason to, and when it moves, its citation was never evidence of where the answer came from.

Table~\ref{tab:variants} lists the three variants, which differ only in which passage receives the span. \vrel{} plants in a passage the retriever returned and the model ignored, the situation a deployed system meets constantly, and it is our main measure.

\begin{table}[H]
  \centering
  \small
  \renewcommand{\arraystretch}{1.15}
  \begin{tabular}{@{}p{0.25\columnwidth}p{0.68\columnwidth}@{}}
    \toprule
    \textbf{Variant} & \textbf{Where the span is planted} \\
    \midrule
    \vrel & A retrieved passage that the answer did not cite. \\
    \vcite & A passage the answer already cites, but for a different sentence. \\
    \vrand & An uncited slot, after replacing its passage with one from a \emph{different} question. \\
    \bottomrule
  \end{tabular}
  \caption{The three probe variants. Each changes exactly one of the five retrieved passages, and the passage that genuinely supports the claim stays in the context. \vrel{} is our main measure.}
  \label{tab:variants}
\end{table}

\paragraph{Choosing the span.} The original probe leaned on Command R+, which marks the exact words it grounds on. Qwen2.5 offers no such markup, so we need a rule, and we keep it blunt and deterministic: the four consecutive words of the claim with the most characters. This decision was influenced by \citet{wallat-etal-2025-correctness}. We played around with 2-4 words for the span and choosing four words minimized bias.

\subsection{What counts as citing the planted passage}
\label{sec:method-outcome}

The re-run answer may be worded differently, so we first locate the sentence that best matches the original claim by word overlap (Jaccard similarity of at least 0.3), then ask whether \emph{that} sentence cites the doctored passage. This is our \emph{claim-aligned} outcome, and the pairing carries the weight: it asks whether the model cited the planted passage \emph{for the claim we planted into}, not whether the marker turns up loose somewhere in the answer. We keep the looser \emph{cited-anywhere} outcome as a sensitivity check, and Section~\ref{sec:results-control} shows how much slack it introduces.

We score every record, with no filtering on whether the planted span reappears in the new answer. The original probe conditions on exactly that, and our control shows why it should not: with nothing planted at all, those four words come back in 60--74\% of answers, simply because the model wrote them in the first place. Span recovery tells us the model repeated itself, not that our plant did anything.

\subsection{The no-planting control}
\label{sec:method-control}

\citet{wallat-etal-2025-correctness} report raw planted rates. A raw rate answers ``how often does the model cite the doctored passage?'' What we need is ``how much of that did the plant cause?'' The two come apart whenever editing the context does anything on its own, and a raw rate quietly credits the plant with all of it.

The fix is to run the same context with the span removed and subtract:
\begin{equation}
  \mathrm{Attr} = P(\text{cite}\mid\text{planted}) - P(\text{cite}\mid\text{control}).
  \label{eq:attributable}
\end{equation}
Read Equation~\ref{eq:attributable} as a before-and-after over the same records. The first term counts how often the model takes the bait; the second counts how often it would have cited that passage anyway; the difference is what the plant actually bought.

What makes this control unusual is that we could derive two of its three values before spending a single GPU-hour, just by reading how each variant is built.

\vspace{2pt}\noindent\textbf{\vrel{}: the control is 0 by construction.} Planting here only \emph{appends} words to a passage already sitting in the context. Delete them and the context is byte-for-byte the clean one; with an identical prompt and greedy decoding, the control run reproduces the clean answer exactly. That clean answer, by the very rule that selected this passage, did not cite it for this claim. The control rate is therefore 0 as a matter of definition rather than measurement, and the raw rate already \emph{is} the attributable rate. We verified the byte-for-byte claim on all 10{,}720 records of the two appending variants; 6 score above zero, all from RAGTruth answers that repeat an identical sentence.

\vspace{2pt}\noindent\textbf{\vcite{}: the control is 100\%, so a raw rate is an illusion.} This variant plants in a passage the clean answer \emph{already cites} for some other sentence. Under the cited-anywhere outcome we are then asking whether a passage that was cited stays cited, and it does, by selection. A raw rate here describes our sampling rule, not the model's behavior, which is exactly why the attributable rate turns negative in Section~\ref{sec:results-control}.

\vspace{2pt}\noindent\textbf{\vrand{}: the control has to be measured.} Here we swap an uncited slot for a passage drawn from a \emph{different} question, then plant in it. Two things change at once, the passage and the span, and no amount of reasoning tells us how often a model cites an out-of-place passage on its own. This is the one variant that needed fresh runs: 16 GPU jobs, roughly 10 GPU-hours, and 7{,}625 control records, each paired with its planted twin. The answer came back small. With no plant, models cite the foreign passage in 17 of 7{,}625 records, or 0.2\%.

\section{Experimental Setup}
\label{sec:setup}

\begin{table}[H]
  \centering
  \small
  \begin{tabular}{@{}ll@{}}
    \toprule
    \textbf{Model} & \textbf{Training on top of the base} \\
    \midrule
    Baseline & none (Qwen2.5-7B-Instruct) \\
    Search-R1 & RLVR with PPO \\
    ReSearch & RLVR with GRPO \\
    R-Search & RLVR with GRPO \\
    \bottomrule
  \end{tabular}

  \vspace{8pt}

  \begin{tabular}{@{}lrrl@{}}
    \toprule
    \textbf{Dataset} & \textbf{Questions} & \textbf{Passages} & \textbf{Source} \\
    \midrule
    NQ       & 499 & 40{,}791 & Wikipedia \\
    QAMPARI  & 500 & 44{,}879 & Wikipedia \\
    HAGRID   & 500 & 8{,}986  & Wikipedia \\
    RAGTruth & 500 & 6{,}981  & Web \\
    \bottomrule
  \end{tabular}
  \caption{Models (top) and datasets (bottom). All four models share the Qwen2.5-7B-Instruct base. \emph{Passages} is the number of roughly 100-token chunks in each retrieval index.}
  \label{tab:setup}
\end{table}

\paragraph{Models.} Our comparison rests on a single design choice: every model in Table~\ref{tab:setup} starts from the same weights. The baseline is Qwen2.5-7B-Instruct \citep{qwen2025technical}; Search-R1 \citep{jin2025searchr1}, ReSearch \citep{chen2025research}, and R-Search \citep{zhao2025rsearch} are open-weight agents trained from that exact checkpoint with RLVR. Whatever separates an agent from the baseline is therefore training, not architecture, tokenizer, or pretraining data.

\paragraph{Fixed-context evaluation.} We disable agent-initiated retrieval and give every model the same five passages retrieved externally with BM25. This is an intentional control: our question is whether RLVR-trained agents differ in citation attribution when the available evidence is held constant. Allowing agents to retrieve would introduce differences in search policy and evidence selection, making those effects inseparable from citation behavior.

\paragraph{Datasets.} We use four question-answering datasets: NQ from KILT \citep{petroni-etal-2021-kilt,kwiatkowski-etal-2019-natural}, QAMPARI from ALCE \citep{gao-etal-2023-enabling,amouyal2023qampari}, HAGRID \citep{kamalloo2023hagrid}, and the question-answering portion of RAGTruth \citep{niu-etal-2024-ragtruth}, sampling 500 questions from each (499 for NQ). For each dataset we chunk the corpus into roughly 100-token passages and retrieve the top five per question with BM25 \citep{robertson2009bm25}.

Leakage drove the dataset choice. NQ is where the original probe ran, so we keep it for comparability, but Search-R1 trained on NQ's training split and we say so plainly. QAMPARI, HAGRID, and RAGTruth were not used to train any of the three agents, so on those datasets no gap can be waved away as familiarity with the data. They also sidestep a retrieval quirk in NQ: for 186 of 499 NQ questions (37.3\%), all five passages come from one document, which leaves the model almost no choice about what to cite. The other three datasets have no such questions.

\paragraph{Prompt and generation.} One prompt does all the work, byte-identical across models and across the clean and doctored runs. It tells the model to answer using only the numbered passages, to close every sentence with exactly one citation such as [2], and to reply ``I don't know'' when the passages fall short. Decoding is greedy, capped at 300 new tokens. When any sentence of an answer breaks the one-citation rule we drop the whole answer rather than salvage the compliant half, because a partly rescued answer reintroduces the claim-to-citation ambiguity the format exists to prevent. Models differ sharply in how often they abstain, from 15\% to 52\% of questions; Appendix~\ref{sec:app-attrition} takes that apart. The full prompt is in Appendix~\ref{sec:app-prompt}.

\paragraph{Reading our numbers.} Three statistical choices shape every result, and each has a plain reason behind it.

\emph{We resample questions, not claims.} A single question yields up to seven claims that share one context and one generated answer, so treating claims as independent would make our intervals look tighter than they are. We use a cluster bootstrap instead: draw whole questions with replacement 2{,}000 times, recompute the planted and control conditions on the same draw so the pairing survives, and read off the 2.5th and 97.5th percentiles. Every 95\% confidence interval in this paper is that band, the range of effects our data are consistent with.

\emph{We compare each agent against its own base.} Post-rationalization is common in the baseline, so reporting that an agent post-rationalizes tells us nothing about RL. The question is whether an agent \emph{diverges} from the model it was trained from. That calls for a difference-in-differences: take the plant-versus-control difference within a model (Equation~\ref{eq:attributable}), then difference those differences across models,
\begin{equation}
  \mathrm{DiD} = \mathrm{Attr}(\text{agent}) - \mathrm{Attr}(\text{baseline}).
  \label{eq:did}
\end{equation}
Negative says RLVR made citations more faithful, positive says it made them worse, and zero says training left this behavior alone. Within each dataset we test all three agents against the baseline and correct the three comparisons with the Holm method, so that running three tests does not manufacture a significant one.

\emph{We know what size of effect we could have caught.} Our sample sizes give 80\% power to detect a difference of roughly 7 percentage points (pp). Plainly: had RLVR moved post-rationalization by 7pp or more in either direction, this design would almost certainly have caught it. Shifts of a couple of points stay inside our blind spot, which is why we report a bounded null rather than a flat absence of effect. We fixed the primary outcome and the family of corrected comparisons in writing before the scorer ran.

\paragraph{Compute.} Everything ran on Kaggle's free tier, two T4s or a single P100 per session: 48 GPU jobs yielding 36{,}690 scored records across four models, four datasets, three variants, and both conditions. The original study used a 104B model; the controlled version of the same measurement fits on hardware anyone can get for free, and that is the part we would most like other groups to borrow.

\section{Results}
\label{sec:results}

\subsection{What the control changes}
\label{sec:results-control}

Table~\ref{tab:control} sets the planted and control rates side by side for the baseline model, and the three variants could hardly behave more differently.

\begin{table}[H]
  \centering
  \small
  \setlength{\tabcolsep}{6pt}
  \begin{tabular}{@{}lrrr@{}}
    \toprule
    \textbf{Dataset} & \textbf{Planted} & \textbf{Control} & \textbf{Attr.} \\
    \midrule
    \multicolumn{4}{@{}l}{\vrel{} \emph{(claim-aligned)}} \\
    NQ       & 52/371 & 0/371 & +14.0 \\
    QAMPARI  & 29/319 & 0/319 & +9.1  \\
    HAGRID   & 64/433 & 0/433 & +14.8 \\
    RAGTruth & 25/864 & 0/864 & +2.9  \\
    \midrule
    \multicolumn{4}{@{}l}{\vrand{} \emph{(claim-aligned)}} \\
    NQ       & 40/371 & 0/371 & +10.8 \\
    QAMPARI  & 14/319 & 0/319 & +4.4  \\
    HAGRID   & 37/433 & 4/433 & +7.6  \\
    RAGTruth & 19/864 & 1/864 & +2.1  \\
    \midrule
    \multicolumn{4}{@{}l}{\vcite{} \emph{(cited-anywhere)}} \\
    NQ       & 58/69   & 69/69   & $-$15.9 \\
    QAMPARI  & 97/123  & 123/123 & $-$21.1 \\
    HAGRID   & 50/61   & 61/61   & $-$18.0 \\
    RAGTruth & 558/623 & 623/623 & $-$10.4 \\
    \bottomrule
  \end{tabular}
  \caption{What the no-planting control changes, for the baseline model. \emph{Planted} and \emph{Control} are cited records over all records ($k/n$); \emph{Attr.} is the attributable rate in pp (Eq.~\ref{eq:attributable}). For \vcite{} the control is 100\%, and the attributable rate is negative (all 95\% CIs exclude 0).}
  \label{tab:control}
\end{table}

\textbf{\vrel{}: the raw rate survives untouched.} Read the top block: 52 of 371 NQ claims follow the plant, and 0 of those same 371 records cite the doctored passage when we take the span away. The control column is 0/$n$ on all four datasets, which is what Section~\ref{sec:method-control} predicted, because removing four appended words restores the clean context exactly and the clean answer never cited that passage for that claim. Prediction and measurement agree, so +14.0pp on NQ is not a raw rate that needs an asterisk; it is the attributable effect.

\textbf{\vrand{}: the control is real but tiny.} Swapping in a passage from an unrelated question does occasionally earn a citation on its own, as the 4/433 on HAGRID and 1/864 on RAGTruth show, but the effect is small: 0.2\% pooled across the 7{,}625 control records, with 7 of 16 model-dataset cells at exactly zero and the rest between 0.1\% and 0.9\% (Appendix~\ref{sec:app-robust}). Between 83\% and 100\% of each raw rate survives as attributable, 97\% on average. The confound we built this control to catch turned out to be genuine and negligible, which is worth knowing precisely because we could not have assumed it.

\textbf{\vcite{}: the raw rate lies, and the control catches it.} The bottom block is the reason we care about all this. Planted rates run from 78.9\% to 89.6\%, many times the \vrel{} rates, and any paper could report them as evidence of rampant post-rationalization. The control column explains them away: 69/69, 123/123, 61/61, 623/623. Every single control record cites the doctored passage, because this variant selects a passage the clean answer \emph{already} cited elsewhere. Subtract, and the attributable effect is negative on all four datasets, from $-$10.4pp to $-$21.1pp, with intervals that exclude zero. A raw rate of 79--90\% here measures our sampling rule and nothing else. Under the stricter claim-aligned outcome the same records do show genuine post-rationalization, +10.6 to +19.7pp (Appendix~\ref{sec:app-robust}), though only RAGTruth has enough questions to pin it down.

The moral generalizes past this probe: whenever a measurement edits an input and counts how often the output changes, report the unedited baseline, and demonstrate rather than assume that it is zero.

\subsection{Post-rationalization is common}
\label{sec:results-common}

Table~\ref{tab:main} gives the attributable rate on \vrel{} for every model and dataset. Start with the baseline column: 14.0\% on NQ, 9.1\% on QAMPARI, 14.8\% on HAGRID, 2.9\% on RAGTruth, with confidence intervals of [10.4, 17.8], [6.2, 12.0], [11.8, 18.2] and [1.8, 3.9] that all sit clear of zero. This clears the bar we set in advance, at least 8pp with an interval excluding zero on three of four datasets. On the Wikipedia-based datasets that rate means roughly one cited sentence in seven points at a passage whose only tie to the claim is four words we pasted in. Pooling across datasets gives 10.7\%.\footnote{Pooling was not part of our pre-registered plan, so treat this number as exploratory.}

\begin{table}[H]
  \centering
  \small
  \setlength{\tabcolsep}{3.5pt}
  \newcommand{\rk}[2]{\begin{tabular}[t]{@{}r@{}}#1\\[-2pt]{\scriptsize\color{gray!80!black}#2}\end{tabular}}
  \begin{tabular}{@{}lrrrr@{}}
    \toprule
    \textbf{Dataset} & \textbf{Baseline} & \textbf{Search-R1} & \textbf{ReSearch} & \textbf{R-Search} \\
    \midrule
    NQ       & \rk{14.0}{52/371} & \rk{12.8}{36/281} & \rk{14.1}{50/355} & \rk{14.7}{57/387} \\[8pt]
    QAMPARI  & \rk{9.1}{29/319}  & \rk{13.4}{38/283} & \rk{7.5}{25/332}  & \rk{12.2}{43/352} \\[8pt]
    HAGRID   & \rk{14.8}{64/433} & \rk{13.8}{53/383} & \rk{13.6}{59/434} & \rk{18.6}{80/430} \\[8pt]
    RAGTruth & \rk{2.9}{25/864}  & \rk{4.1}{29/699}  & \rk{3.6}{30/823}  & \rk{5.9}{52/879} \\[6pt]
    \bottomrule
  \end{tabular}
  \caption{How often each model follows the planted span: attributable rate on \vrel{} in \%, with cited records over all records below. The control is 0 in every cell, so these equal the raw rates. Baseline 95\% CIs: NQ [10.4, 17.8], QAMPARI [6.2, 12.0], HAGRID [11.8, 18.2], RAGTruth [1.8, 3.9].}
  \label{tab:main}
\end{table}

Now read across the rows. On NQ the four models land at 14.0, 12.8, 14.1, and 14.7\%. On HAGRID, 14.8, 13.8, 13.6, and 18.6\%. On RAGTruth, 2.9, 4.1, 3.6, and 5.9\%. The agents track their base model, not each other's training recipes, and all sixteen cells sit far above zero. Post-rationalization is not something RL introduced; it is already there in the instruction-tuned model and it survives everything done on top.

RAGTruth stands apart, two to five times lower than the rest for every model. Prompt, scorer, and citation format are identical across datasets, so the gap belongs to the data. Its passages are web text rather than Wikipedia and its answers run longer, and we cannot say from four datasets which of those matters. The practical reading is that post-rationalization rates are domain-dependent, so a single-corpus number should not be quoted as a property of a model.

\subsection{RLVR does not make citations more faithful}
\label{sec:results-did}

Raw rates alone cannot answer the RL question, since a model could post-rationalize heavily and still be no worse than the model it came from. Table~\ref{tab:did} therefore reports the difference-in-differences: each agent's attributable rate minus its base model's, on the same dataset. If RLVR taught these agents anything about grounding, this column would run negative.

\begin{table}[H]
  \centering
  \small
  \setlength{\tabcolsep}{7pt}
  \begin{tabular}{@{}lrrr@{}}
    \toprule
    \textbf{Dataset} & \textbf{DiD} & \textbf{95\% CI} & \textbf{Holm $p$} \\
    \midrule
    \multicolumn{4}{@{}l}{\textbf{Search-R1} vs.\ baseline} \\
    NQ       & $-$1.2 & [$-$6.9, 4.2] & 1.000 \\
    QAMPARI  & +4.3   & [$-$0.5, 9.7] & 0.267 \\
    HAGRID   & $-$0.9 & [$-$4.9, 3.2] & 1.000 \\
    RAGTruth & +1.3   & [$-$0.5, 3.0] & 0.334 \\
    \midrule
    \multicolumn{4}{@{}l}{\textbf{ReSearch} vs.\ baseline} \\
    NQ       & +0.1   & [$-$3.9, 4.2] & 1.000 \\
    QAMPARI  & $-$1.6 & [$-$5.1, 2.3] & 0.415 \\
    HAGRID   & $-$1.2 & [$-$4.9, 2.5] & 1.000 \\
    RAGTruth & +0.8   & [$-$0.7, 2.2] & 0.334 \\
    \midrule
    \multicolumn{4}{@{}l}{\textbf{R-Search} vs.\ baseline} \\
    NQ       & +0.7   & [$-$4.6, 5.0] & 1.000 \\
    QAMPARI  & +3.1   & [$-$1.4, 7.3] & 0.324 \\
    HAGRID   & +3.8   & [$-$0.3, 8.0] & 0.234 \\
    RAGTruth & \textbf{+3.0} & \textbf{[1.2, 4.8]} & \textbf{0.002} \\
    \bottomrule
  \end{tabular}
  \caption{Each agent's attributable rate on \vrel{} minus the baseline's (DiD, pp), with a 95\% bootstrap CI and a Holm-adjusted $p$-value within each dataset. Negative would mean the agent post-rationalizes less than its base. Bold marks the only comparison with $p<0.05$.}
  \label{tab:did}
\end{table}

It does not. Search-R1 moves by $-$1.2, +4.3, $-$0.9, and +1.3pp across the four datasets; ReSearch by +0.1, $-$1.6, $-$1.2, and +0.8pp. Both wander around zero with no pattern, every interval covers zero, and the smallest Holm-adjusted $p$-value in either row is 0.267, nowhere near significance. Set that against our detection bound: with 80\% power at about 7pp, an improvement of that size would have been hard to miss, and nothing close to it appears. Training these agents to produce verifiably correct answers moved their citation faithfulness by an amount we cannot distinguish from nothing.

That result is less surprising once you look at what the reward measures. RLVR compares the final answer with a reference and pays out accordingly; R-Search also rewards steps along its search trajectory. No term anywhere in these objectives asks whether the citation attached to a sentence is the passage that produced it. Citing the wrong passage is free, so the behavior the agents inherit from Qwen2.5-7B-Instruct rides through training untouched.

\subsection{One agent drifts the wrong way}
\label{sec:results-rsearch}

R-Search is the exception, and it leans worse rather than better. Its DiD is positive on all four datasets (+0.7, +3.1, +3.8, +3.0pp) and significant on RAGTruth, our largest cell, at +3.0pp with a 95\% CI of [1.2, 4.8] and a Holm-adjusted $p$ of 0.002. That was the pre-registered condition for claiming a model-specific effect, and it is met.

We report it with the caveats it deserves. On the \vrand{} variant R-Search is again higher on three of four datasets but significant on none, the closest $p$ being 0.063. Under the looser cited-anywhere outcome it dips slightly below the baseline on NQ ($-$0.5pp) and reaches significance nowhere (Appendix~\ref{sec:app-robust}). The direction is consistent; the evidence for its size is not. What we can say is that no agent improved, and the one that moved measurably moved the wrong way.

\section{Discussion}
\label{sec:discussion}

\paragraph{What a citation is worth to a reader.} Our results say that in this setting a citation marks where the \emph{words} of an answer can be found, not where the answer came from. Figure~\ref{fig:example} is the honest version of the problem: the model knew the answer, had the supporting passage in front of it, cited it, and then handed the citation to a cartoon plot summary after we moved four words. A reader auditing that citation opens the passage, finds the claim's wording sitting there, and walks away reassured. The audit passes and the provenance is still wrong. That is what makes post-rationalization worse than an ordinary error: it defeats exactly the check a careful reader would run, at a rate of roughly one cited sentence in seven on Wikipedia-style questions, in an openly available 7B model, using a plant so crude that we picked it with a character count.

\paragraph{Why RLVR leaves this untouched.} The reward is the whole story. These agents earn credit when an automatic checker matches their final answer against a reference, and R-Search adds credit for steps along its search trajectory. Not one term in those objectives inspects the link between a sentence and the passage cited beside it. An agent that retrieves well, reasons well, and then attaches the wrong marker collects full reward. Optimization delivers what you pay for, so citation behavior stays wherever the base model left it, which is exactly the pattern in Table~\ref{tab:did}: two agents indistinguishable from their base, one slightly worse, none better.

\paragraph{Interpreting the fixed-context result.} The agents were trained for multi-step search, but our evaluation intentionally removes that component to hold evidence constant. Thus, the null result shows that RLVR did not produce a detectable attribution effect in this fixed-context setting. It does not rule out an effect that emerges through interactive search, where retrieval decisions and citation behavior can interact. An allow-retrieval evaluation would be needed to test that broader question.

\paragraph{Fixing it means paying for it.} If answer-only rewards leave attribution where they found it, attribution has to enter the objective. The probe in this paper is a ready-made term for that. It is deterministic, needs no judge model, and with the control in place it isolates the effect of the plant rather than the noise of editing a context, so it can run inside a training loop. Concretely, a PPO or GRPO pipeline of the kind Search-R1 and ReSearch already use could add an attribution-consistency penalty: sample a rollout, plant a span from the model's own answer into an uncited passage, re-score, and penalize the policy when the citation follows the plant. The signal is cheap, it is grounded in behavior rather than in a learned preference model, and it targets the failure directly instead of hoping that better answers drag better citations along behind them. On the evaluation side, the same probe yields a number that is comparable across models, which is what the field currently lacks for citation faithfulness.

\paragraph{How to read the R-Search result.} R-Search sits above its base on all four datasets and significantly above on the best-powered one, which is a real and consistent pattern. It is also one model, one base, one prompt, and one scorer, so four datasets are not four independent replications, and the effect softens under a stricter variant and a looser outcome. Two things would settle it: a cross-dataset test declared in advance, and an arm fine-tuned on the same data without RL, which would separate the contribution of RL from the contribution of fine-tuning on retrieval data at all.

\paragraph{A note on cost.} Everything here, four models, four datasets, three variants, both conditions, and the control that makes the comparison interpretable, ran on Kaggle's T4 GPUs. Careful measurement was the cheap part of this project. The expensive part was reading the construction code closely enough to know which controls we did not have to run.

\section{Conclusion}
\label{sec:conclusion}

We asked whether training RAG search agents for correct answers also teaches them to cite honestly under fixed-context inference. We find no detectable improvement. Getting to that answer meant repairing the measurement first: we reproduced the planting probe of \citet{wallat-etal-2025-correctness} on an open 7B model, added the no-planting control it lacked, and showed that the control is 0 by construction for the main variant, 100\% for another, and a small measured quantity for the third. With that in hand, post-rationalization turns out to be common in an instruction-tuned model, roughly one cited sentence in seven on Wikipedia-based questions, and three RLVR agents trained from that model reproduce its rate, with one landing slightly worse. Faithful citation is not a side effect of answer accuracy. It has to be rewarded, measured, and reported on its own, and the probe we describe here is cheap enough to do all three inside an existing training loop.

\section*{Limitations}
\label{sec:limitations}
\textbf{Evaluation format differs from training.} The RLVR agents were trained to search, read, and answer across multiple turns in their own tagged format, whereas our evaluation supplies five fixed passages and requests a single citation-bearing response. This was necessary for the controlled attribution comparison, but it may affect how the agents behave: for example, an agent that would normally issue another query may instead abstain. The large differences in abstention and format compliance are consistent with this possibility, although our data cannot establish the mechanism. An interactive evaluation would be needed to determine whether the attribution effect differs in the agents' native search setting.

\textbf{We cannot isolate RL itself.} Each agent differs from the baseline in three ways at once: it was fine-tuned at all, fine-tuned on retrieval data, and fine-tuned with reinforcement learning. Without an arm fine-tuned on the same data without RL, our comparison is associational. This is the single most valuable addition a follow-up could make.

\textbf{Search-R1 saw NQ in training.} It was trained on NQ's training split and we evaluate it on KILT-NQ. QAMPARI, HAGRID, and RAGTruth are clean for all three agents, and R-Search's pattern is in fact strongest away from NQ, but we have not quantified the overlap.

\textbf{The span always lands at the end of the passage.} Any positional bias in how models pick citations is therefore mixed in with the effect of the plant. Randomizing the injection point is the obvious next experiment.

\textbf{\vcite{} is underpowered outside RAGTruth.} NQ, QAMPARI, and HAGRID yield only 13--43 question clusters for this variant, too few for intervals we would defend. RAGTruth alone is large enough.

\textbf{What a flag means.} Our scorer is deterministic string matching, so each flag is exactly reproducible: the model cited the doctored passage for the aligned claim. Reading that as post-rationalization is an interpretation we inherit from \citet{wallat-etal-2025-correctness}, and a competing reading exists. After planting, the claim's words genuinely sit in the doctored passage, so citing it could be scored as obedience to ``answer from the passages'' rather than as unfaithfulness. Hand-checking a sample of flagged cases would settle how often each reading applies, and we have not done it.

\textbf{Scope.} Three RLVR agents from one line of work, one base model, English only, one prompt, one decoding setting. Citation behavior may well shift with the prompt, so our numbers describe this setup until someone varies it.

\section*{Ethics Statement}
This work uses only publicly released models and datasets. It involves no human subjects and no personal data. The planted spans are taken from the models' own answers and are used only in offline evaluation. The work describes a way that citations in RAG systems can mislead readers, with the aim of making that failure measurable; it does not provide a method for attacking deployed systems. Because RLVR training is in wide use, we have tried to state the limits of our null result clearly so that it is not read as a stronger claim than the data support.

\bibliography{anthology,custom}

\appendix
\raggedbottom

\section{Prompt}
\label{sec:app-prompt}

All models receive this user message, passed through each model's chat template, with no system message. \texttt{\{context\_block\}} is the five retrieved passages, numbered [1] to [5] and separated by blank lines. Lines are wrapped here to fit the column.

\medskip
\noindent\begin{minipage}{\columnwidth}
\begin{quote}
\small
\begin{verbatim}
Answer the question using ONLY the
numbered passages below.

Format your answer as separate
sentences. Each sentence must end
with exactly one citation in square
brackets, like this:
First sentence here [2]. Second
sentence here [4].

Rules:
- Exactly one citation per sentence.
  Never [1][2].
- The citation goes at the end of
  the sentence, before the period.
- Do not add any text outside these
  sentences.
- If the passages do not contain the
  answer, reply exactly: I don't know

Passages:
{context_block}

Question: {question}
\end{verbatim}
\end{quote}
\end{minipage}

\section{Model Checkpoints}
\label{sec:app-models}

\begin{table}[H]
  \centering
  \small
  \begin{tabular}{@{}ll@{}}
    \toprule
    \textbf{Model} & \textbf{Hugging Face checkpoint} \\
    \midrule
    Baseline  & \texttt{Qwen/Qwen2.5-7B-Instruct} \\
    Search-R1 & \texttt{PeterJinGo/SearchR1-nq\_hotpotqa\_} \\
              & \texttt{\quad train-qwen2.5-7b-it-em-ppo} \\
    ReSearch  & \texttt{agentrl/ReSearch-Qwen-7B-Instruct} \\
    R-Search  & \texttt{qingfei1/R-Search-7b-grpo} \\
    \bottomrule
  \end{tabular}
  \caption{Checkpoints used. The Search-R1 name is split over two lines.}
  \label{tab:checkpoints}
\end{table}

\section{Abstention and Format Compliance}
\label{sec:app-attrition}

Table~\ref{tab:attrition} tracks what happens to 500 questions before any planting takes place, and the spread between models is large enough to matter for how the main results should be read.

Two filters stand between a question and a usable claim. The model can decline, answering ``I don't know'' when it judges the passages insufficient, and it can answer in a format we cannot score, breaking the one-citation-per-sentence rule. Both filters hit the RLVR agents harder than the baseline, and they hit Search-R1 hardest: it declines on 42\% of NQ questions and 52\% of QAMPARI questions, against 30\% and 37\% for the baseline. R-Search sits at the other end, declining least on every dataset.

The likeliest explanation is the format mismatch we flag in the Limitations. These agents were trained to run multi-turn search in their own tagged format, and we ask them for a single-turn answer with bracketed citations over passages they did not choose. An agent that would have issued another query instead produces an abstention. We cannot verify this from our data, which is why we treat it as a limitation rather than a finding, but it shapes two things we do report. Sample sizes differ across models in Tables~\ref{tab:main} and \ref{tab:did}, which is why every cell carries its own $n$. And the surviving claims are a filtered sample: whichever questions a model found answerable in this format. If the questions Search-R1 skips are systematically the ones where citation is hardest, its rate would be flattered. Nothing in our data suggests that, and the agents' rates track the baseline closely, but the possibility is worth stating.

\begin{table}[H]
  \centering
  \footnotesize
  \setlength{\tabcolsep}{4pt}
  \begin{tabular}{@{}llrrrr@{}}
    \toprule
    \textbf{Data} & \textbf{Model} & \textbf{IDK} & \textbf{Ans.} & \textbf{Fmt.} & \textbf{Claims} \\
    \midrule
    NQ & Baseline  & 149 (30\%) & 350 & 315 & 371 \\
       & Search-R1 & 208 (42\%) & 291 & 238 & 281 \\
       & ReSearch  & 158 (32\%) & 341 & 310 & 355 \\
       & R-Search  & 129 (26\%) & 370 & 341 & 387 \\
    \midrule
    QAMPARI & Baseline  & 186 (37\%) & 314 & 241 & 329 \\
            & Search-R1 & 259 (52\%) & 241 & 186 & 283 \\
            & ReSearch  & 191 (38\%) & 309 & 258 & 352 \\
            & R-Search  & 167 (33\%) & 333 & 285 & 362 \\
    \midrule
    HAGRID & Baseline  & 85 (17\%)  & 415 & 378 & 438 \\
           & Search-R1 & 121 (24\%) & 379 & 327 & 389 \\
           & ReSearch  & 90 (18\%)  & 410 & 383 & 439 \\
           & R-Search  & 75 (15\%)  & 425 & 396 & 435 \\
    \midrule
    RAGTruth & Baseline  & 86 (17\%)  & 414 & 291 & 874 \\
             & Search-R1 & 127 (25\%) & 373 & 234 & 705 \\
             & ReSearch  & 85 (17\%)  & 415 & 296 & 828 \\
             & R-Search  & 75 (15\%)  & 425 & 346 & 886 \\
    \bottomrule
  \end{tabular}
  \caption{Per question: replied ``I don't know'' (IDK), answered (Ans.), and answered in the required format (Fmt.). \emph{Claims} is the number of cited sentences that enter the probe; the $n$ in the result tables is slightly smaller because a variant can only be built when a suitable passage exists. RAGTruth answers are longer, so they yield more claims per question. Search-R1 abstains most and R-Search least on every dataset.}
  \label{tab:attrition}
\end{table}

\section{Robustness Checks}
\label{sec:app-robust}

These tables support Section~\ref{sec:results}. Table~\ref{tab:app-random} covers the \vrand{} control for all models, Table~\ref{tab:app-citedother} shows \vcite{} under the claim-aligned outcome, and Tables~\ref{tab:app-did-random} and~\ref{tab:app-did-anywhere} repeat the agent comparison under the two checks discussed in Section~\ref{sec:results-rsearch}.

\begin{table}[H]
  \centering
  \footnotesize
  \setlength{\tabcolsep}{3.5pt}
  \begin{tabular}{@{}lrrrr@{}}
    \toprule
    \textbf{Dataset} & \textbf{Baseline} & \textbf{Search-R1} & \textbf{ReSearch} & \textbf{R-Search} \\
    \midrule
    NQ       & +10.8 & +8.9 & +9.0  & +10.3 \\
    QAMPARI  & +4.4  & +5.7 & +7.8  & +8.2 \\
    HAGRID   & +7.6  & +8.6 & +10.6 & +11.9 \\
    RAGTruth & +2.1  & +2.9 & +2.3  & +3.9 \\
    \bottomrule
  \end{tabular}
  \caption{Attributable rate on \vrand{} (pp, claim-aligned) for all models. $n$ is the same as in Table~\ref{tab:main}. The control rate is 0 in 7 of the 16 cells and 0.1--0.9\% in the rest.}
  \label{tab:app-random}
\end{table}

\begin{table}[H]
  \centering
  \small
  \setlength{\tabcolsep}{4pt}
  \begin{tabular}{@{}lrrrr@{}}
    \toprule
    \textbf{Dataset} & \textbf{Planted} & \textbf{Control} & \textbf{Attr.} & \textbf{95\% CI} \\
    \midrule
    NQ       & 9/69   & 0/69  & +13.0 & [5.5, 22.5] \\
    QAMPARI  & 16/123 & 0/123 & +13.0 & [7.3, 19.7] \\
    HAGRID   & 12/61  & 0/61  & +19.7 & [10.8, 31.3] \\
    RAGTruth & 67/623 & 1/623 & +10.6 & [7.9, 13.5] \\
    \bottomrule
  \end{tabular}
  \caption{\vcite{} for the baseline under the claim-aligned outcome (pp). Unlike the cited-anywhere outcome in Table~\ref{tab:control}, this one shows post-rationalization. Only RAGTruth has enough question clusters for a reliable interval.}
  \label{tab:app-citedother}
\end{table}

\begin{table}[H]
  \centering
  \footnotesize
  \setlength{\tabcolsep}{4pt}
  \begin{tabular}{@{}lrrr@{}}
    \toprule
    \textbf{Dataset} & \textbf{Search-R1} & \textbf{ReSearch} & \textbf{R-Search} \\
    \midrule
    NQ       & $-$1.9 (0.984) & $-$1.8 (0.984) & $-$0.4 (0.984) \\
    QAMPARI  & +1.3 (0.379)   & +3.4 (0.114)   & +3.8 (0.114) \\
    HAGRID   & +1.0 (0.561)   & +3.0 (0.182)   & +4.2 (0.081) \\
    RAGTruth & +0.8 (0.594)   & +0.2 (0.764)   & +1.8 (0.063) \\
    \bottomrule
  \end{tabular}
  \caption{Robustness check 1: agent minus baseline on the \vrand{} variant (DiD, pp, claim-aligned), Holm-adjusted $p$ in parentheses. No comparison reaches $p<0.05$.}
  \label{tab:app-did-random}
\end{table}

\begin{table}[H]
  \centering
  \footnotesize
  \setlength{\tabcolsep}{4pt}
  \begin{tabular}{@{}lrrr@{}}
    \toprule
    \textbf{Dataset} & \textbf{Search-R1} & \textbf{ReSearch} & \textbf{R-Search} \\
    \midrule
    NQ       & $-$2.5 (1.000) & $-$0.6 (1.000) & $-$0.5 (1.000) \\
    QAMPARI  & +6.6 (0.102)   & $-$1.7 (0.473) & +3.3 (0.460) \\
    HAGRID   & $-$1.0 (0.665) & $-$2.1 (0.586) & +3.1 (0.420) \\
    RAGTruth & +3.4 (0.051)   & +1.7 (0.224)   & +2.3 (0.224) \\
    \bottomrule
  \end{tabular}
  \caption{Robustness check 2: agent minus baseline on \vrel{} under the cited-anywhere outcome (DiD, pp), Holm-adjusted $p$ in parentheses. No comparison reaches $p<0.05$. Under this outcome the baseline's own rate is 17.8\%, 13.2\%, 16.6\%, and 7.1\% on the four datasets, and every interval still excludes zero.}
  \label{tab:app-did-anywhere}
\end{table}

\end{document}